\documentclass[runningheads]{llncs}
\usepackage{graphicx}
\usepackage{comment}
\usepackage{marvosym}
\usepackage{amsmath}
\usepackage{amssymb}

\newcommand{\ie}{\textit{i}.\textit{e}.}

\usepackage{numprint}
\usepackage{color}
\usepackage{url}
\usepackage{multirow}
\usepackage{wrapfig}
\usepackage{hyperref}
\usepackage[format=plain,labelformat=simple,labelsep=period,labelfont=bf,font=small,skip=6pt,compatibility=false]{caption}
\usepackage[font=footnotesize,skip=3pt]{subcaption}
\usepackage{booktabs,tabularx}
\usepackage[inline]{enumitem}
\usepackage[capitalize]{cleveref}
\usepackage{tikz}
\newcommand{\eg}{\textit{e}.\textit{g}.}
\newcommand{\cf}{\textit{cf}.}

\usepackage{balance}
\usepackage{fancyhdr}

\newcommand\blfootnote[1]{%
  \begingroup
  \renewcommand\thefootnote{}\footnote{#1}%
  \addtocounter{footnote}{-1}%
  \endgroup
}

\newcommand{\myparagraph}[1]{\paragraph{#1}}

\newif\ifreview
\reviewfalse

\ifreview
	\usepackage{lineno}

	\linenumbers
\fi

\begin{document}


\def\SubNumber{075}

\def\GCPRTrack{Regular Track}

\title{Diverse Image Captioning with Grounded Style}

\ifreview
\else

	\author{Franz Klein\inst{1} \and
	Shweta Mahajan\inst{1} \and
	Stefan Roth\inst{1,2}}
	
	\authorrunning{F.~Klein et al.}
	
	\institute{Department of Computer Science, TU Darmstadt, Darmstadt, Germany \\
	 \and
	hessian.AI, Darmstadt, Germany}
\fi

\maketitle              
\thispagestyle{fancy}

\begin{abstract}
Stylized image captioning as presented in prior work aims to generate captions that reflect characteristics beyond a factual description of the scene composition, such as sentiments.
Such prior work relies on \emph{given} sentiment identifiers, which are used to express a certain global style in the caption, \eg~positive or negative, however without taking into account the stylistic content of the visual scene.
To address this shortcoming, we first analyze the limitations of current stylized captioning datasets and propose COCO attribute-based augmentations to obtain varied stylized captions from COCO annotations.
Furthermore, we encode the stylized information in the latent space of a Variational Autoencoder; specifically, we leverage extracted image attributes to explicitly structure its sequential latent space according to different localized style characteristics.
Our experiments on the Senticap and COCO datasets show the ability of our approach to generate accurate captions with diversity in styles that are grounded in the image. \blfootnote{Code available at: \url{https://github.com/visinf/style-seqcvae}}

\end{abstract}
\section{Introduction}
Recent advances in deep learning and the availability of multi-modal datasets at the intersection of vision and language \cite{Lin2014,flickr} have led to the successful development of image captioning models \cite{Anderson2017,image_captioning_gan,mahajan2019joint,mahajan2020diverse,Rennie2016}.
Most of the available datasets for image captioning, \eg~COCO \cite{Lin2014}, consist of several ground-truth captions per image from different human annotators, each of which factually describes the scene composition.
In general, captioning frameworks leveraging such datasets deterministically generate a single caption per image \cite{AnejaDS18,ChenZ15,JohnsonKF16,KarpathyF17,KulkarniPODLCBB13,Lu2018,mrnn,YaoPL019}. 
However, it is generally not possible to express the entire content of an image in a single, human-sounding sentence.
Diverse image captioning aims to address this limitation with frameworks that are able to generate several \emph{different} captions for a single image \cite{Aneja2019,mahajan2020latent,Wang2017}. 
Nevertheless, these approaches largely ignore image and text properties that go beyond reflecting the scene composition; in fact, most of the employed training datasets hardly consider such properties.

Stylized image captioning summarizes these properties under the term \emph{style}, which includes variations in linguistic style through variations in language, choice of words and sentence structure, expressing different emotions about the visual scene, or by paying more attention to one or more localized concepts, \eg~attributes associated with objects in the image \cite{Nezami2018}.
To fully understand and reproduce the information in an image, it is inevitable to consider these kinds of characteristics.
Existing image captioning approaches  \cite{Guo2019,Mathews2015,Nezami2018} implement style as a global sentiment and strictly distinguish the sentiments into `positive', `negative', and sometimes `neutral' categories. 
This simplification ignores characteristics of styles that are crucial for the comprehensive understanding and reproduction of visual scenes. 
Moreover, they are designed to produce one caption based on a given sentiment identifier related to one sentiment category, ignoring the actual stylistic content of the corresponding image \cite{Karayil2019,Mathews2015,Nezami2018}.

In this work, we attempt \emph{(1)} to obtain a more diverse representation of style, and \emph{(2)} ground this style in attributes from localized image regions.
We propose a Variational Autoencoder (VAE) based framework, Style-SeqCVAE, to generate stylized captions with styles expressed in the corresponding image.
To this end, we address the lack of image-based style information in existing captioning datasets \cite{KarpathyF17,Mathews2015} by extending the ground-truth captions of the COCO dataset \cite{KarpathyF17}, which focus on the scene composition, with localized attribute information from different image regions of the visual scene \cite{Patterson2016}.
This style information in the form of diverse attributes is encoded in the latent space of the Style-SeqCVAE. 
We perform extensive experiments to show that our approach can indeed generate captions with image-specific stylized information with high semantic accuracy \emph{and} diversity in stylistic expressions.

\section{Related Work}
\paragraph{Image captioning.} 
A large proportion of image captioning models rely on Long Short-Term Memories (LSTMs) \cite{HochreiterS97} as basis for language modeling in an encoder-decoder or compositional architecture \cite{DonahueHRVGSD17,KarpathyF17,Lu2016,mrnn,VinyalsTBE15,Xu2015}.
These methods are designed to generate a single accurate caption and, therefore, cannot model the variations in stylized content for an image.
Deep generative model-based architectures \cite{Aneja2019,image_captioning_gan,mahajan2020latent,mahajan2020diverse,Wang2017} aim to generate multiple captions, modeling feature variations in a low-dimensional space.
Wang et al.~\cite{Wang2017} formulate constrained latent spaces based on object classes in a Conditional Variational Autoencoder (CVAE) framework. 
Aneja et al.~\cite{Aneja2019} use a sequential latent space to model captions with Gaussian priors and Mahajan et al.~\cite{mahajan2020latent} learn domain-specific variations of images and text in the latent space. 
All these approaches, however, do not allow to directly control the intended style to be reflected in each of the generated captions. This is crucial to generate captions with stylistic variation grounded in the images.
In contrast, here we aim to generate diverse captions with the many localized styles representative of the image.

\myparagraph{Stylized image captioning.}
Recent work has considered the task of stylized caption generation \cite{Chen2018b,Mathews2015,Shin2016,You2018}.
Mathews et al.~\cite{Mathews2015} utilize 2000 stylized training captions in addition to those available from the COCO dataset to generate captions with either positive or negative sentiments.
Shin et al.~\cite{Shin2016} incorporate an additional CNN, solely trained on weakly supervised sentiment annotations of a large image corpus scraped from different platforms.
These approaches, however, rely on \emph{given} style indicators during inference to generate new captions, to emphasize sentiment words in stylized captions, which may not reflect the true sentiment of the image. 
A series of works attempts to overcome the lack of available stylized captioning training data by separating the style components and the remaining information from the textual description such that they can be trained on both factual caption-image pairs and a distinct stylized text corpus.
Gan et al.~\cite{Gan2017} share components in the LSTM over sentences of a particular style. 
Similarly, Chen et al.~\cite{Chen2018b} use self-attention to adaptively consider semantics or style in each time step. 
You et al.~\cite{You2018} add a sentiment cell to the LSTM and 
Nezami et al.~\cite{Nezami2018} apply the well established semantic attention idea to stylized captioning. 
Various approaches utilize adversarial formulations to generate stylized captions \cite{Guo2019,Karayil2019,Nezami2019}.
However, they also rely on given globally positive or negative sentiment identifiers to generate varied captions.
Since an image can contain a variety of localized styles, in this work, we instead focus on generating diverse captions where the style information is \emph{locally grounded in the image}.

\begin{figure}[t]
\begin{subfigure}[b]{0.5\textwidth}
\includegraphics[width=\linewidth]{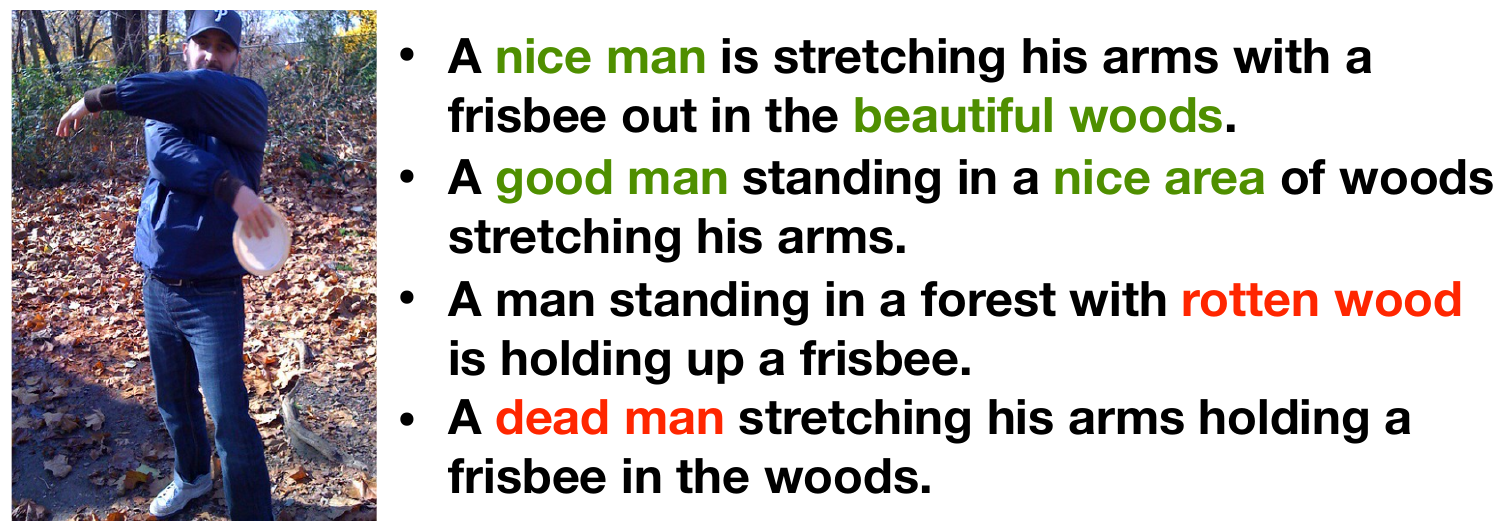}
\caption{Example from the Senticap dataset.}
\label{fig:relwork_senticap_example}
\end{subfigure}
\hfill
\begin{subfigure}[b]{0.3\textwidth}
\includegraphics[width=\linewidth]{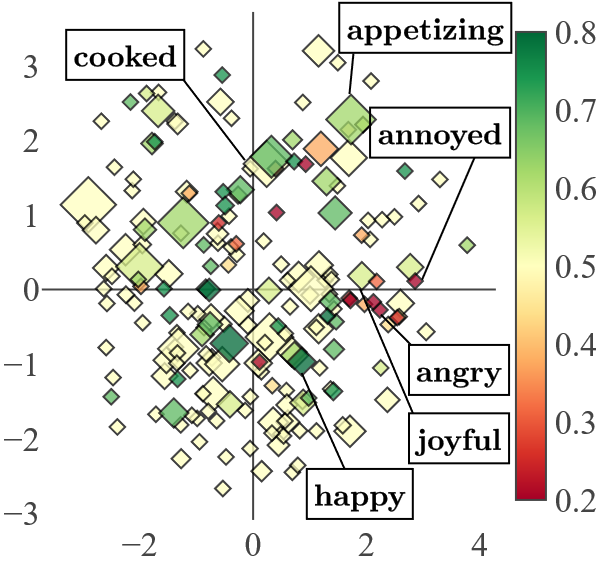}
\caption{GloVe vectors of attributes from COCO Attributes \cite{Patterson2016}.}
\label{fig:methods_cocoatts_distr}
\end{subfigure}
\caption{\emph{(a)} Senticap example ground-truth captions with globally positive or negative sentiment. \emph{(b)}~GloVe vectors of captions with adjectives from COCO Attributes. The color indicates the attribute SentiWordNet score \cite{swd} and the scale indicates the frequency of occurrence in the entire dataset. The original SentiWordNet scores are rescaled between 0 for the most negative and 1 for the most positive sentiment.}
\end{figure}

\section{Image Captioning Datasets with Stylized Captions}
We begin by discussing the limitations of existing image captioning datasets with certain style information in the captions, specifically the Senticap dataset \cite{Mathews2015}.
Further, we introduce an attribute (adjective) insertion method to extend the COCO dataset with captions containing different localized styles. 

\myparagraph{Senticap.}
Besides its limited size (1647 images and 4419 captions in the training set), the Senticap dataset \cite{Mathews2015} does not provide a comprehensive impression of the image content in combination with different sentiments anchored in the image (\cref{fig:relwork_senticap_example}).
The main issue in the ground-truth captions is that the positive or negative sentiments may not be related to the sentiment actually expressed in the image. 
Some adjectives may even distort the image semantics. 
For instance, the man shown in \cref{fig:relwork_senticap_example} is anything but dead, though that is exactly what the last ground-truth caption describes.
Another issue is the limited variety: There are 842 positive adjective-noun pairs (ANPs) composed of 98 different adjectives combined with one out of 270 different nouns. 
For the negative set only 468 ANPs exist, based on 117 adjectives and 173 objects. 
These adjectives, in turn, appear with very different frequencies in the caption set, \cf~\cref{fig:methods_senticap_distr}.

\myparagraph{COCO Attributes.} \label{sec:methods_cocoatts}
The attributes of COCO Attributes \cite{Patterson2016} have the big advantage over Senticap that they actually reflect image information. 
Furthermore, the average number of 9 attribute annotations per object may reflect different possible perceptions of object characteristics. 
On the downside, COCO contains fairly neutral, rather positive than negatively connoted images: for instance, many image scenes involve animals, children, or food. 
This is reflected in the sentiment intensity of the associated attributes, visualized in \cref{fig:methods_cocoatts_distr}.
Most of them tend to be neutral or positive; the negative ones are underrepresented. 
%
%
%
%
\subsection{Extending ground-truth captions with diverse style attributes} \label{sec:methods_caption_augmentation}
\begin{figure}[t]
\centering
\includegraphics[width=0.8\textwidth]{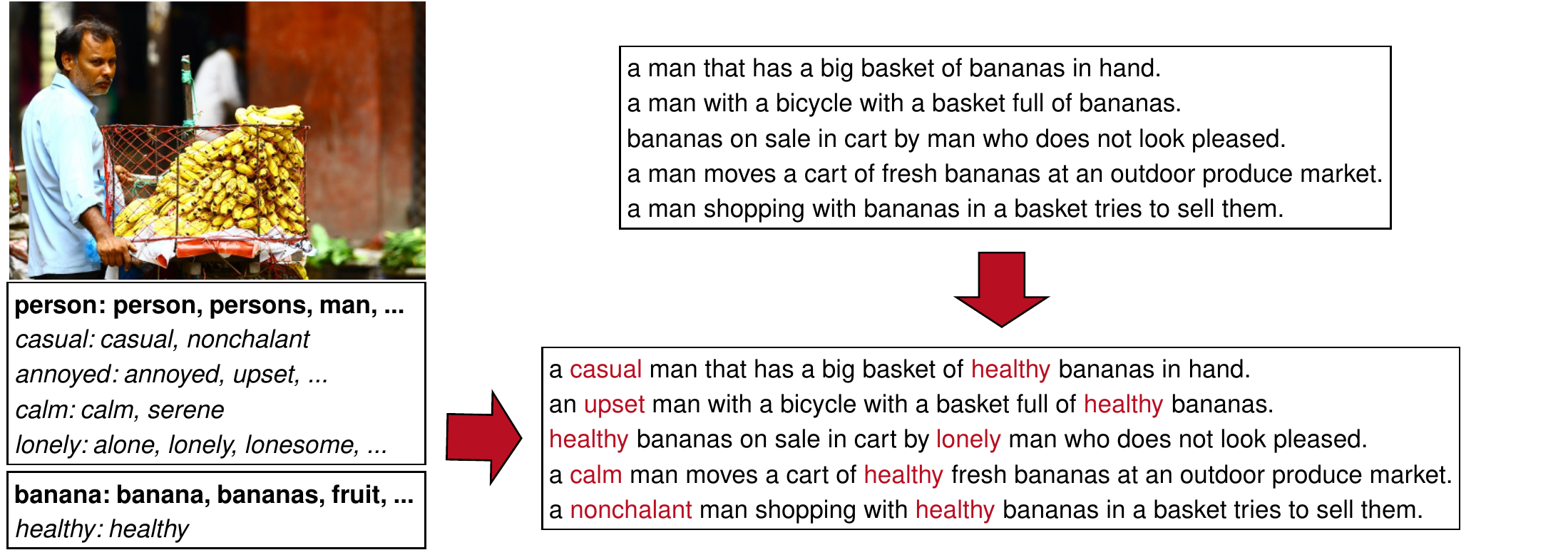}
\caption{Example of COCO caption augmentation by insertion of random samples from the COCO Attributes synonym sets in front of the nouns of COCO object categories. }
\label{fig:methods_augmented_coco}
\end{figure}

To address the lack of stylized ground-truth captions, we combine COCO captions \cite{Lin2014}, focusing on the scene composition, with style-expressive adjectives in COCO Attributes \cite{Patterson2016}. 
We remove 98 attribute categories that are less relevant for stylized captioning (\eg, ``cooked'') and define sets of synonyms for the remaining attributes to increase diversity. 
Some of the neutral adjective attributes are preserved since recognizing and dealing with object attributes having a neutral sentiment is also necessary to fully solve captioning with image-grounded styles. However, most of the 185 adjectives are either positive or negative.
Likewise for the various COCO object categories, we initially define sets of nouns that appear interchangeably in the corresponding captions to name this object category. Subsequently, we iterate over all COCO images with available COCO Attributes annotations. 
Given a COCO image, associated object/attribute labels, and the existing ground-truth captions, we locate nouns in the captions that also occur in the sets of object categories to insert a sampled adjective of the corresponding attribute annotations in front of it. 
A part-of-speech tagger \cite{nltk} helps to insert the adjective at the right position. 
This does not protect against an adjective being associated with the wrong noun if it occurs multiple times in the same caption. 
However, our observations suggest this to be rare. 
The example of the augmentation in \cref{fig:methods_augmented_coco} illustrates that inserting various adjectives could reflect a certain level of ambiguity in perception.
Based on the COCO 2017 train split \cite{KarpathyF17}, this gives us a set of 266K unique, adjective-augmented and image-grounded training captions, which potentially reflect the image style. 

\subsection{Extending the Senticap dataset}
Since most prior work is evaluated on the Senticap dataset \cite{Karayil2019,Mathews2015,Mathews2018,Nezami2018}, whose characteristics strongly differ from COCO Attributes, using COCO Attributes-augmented captions for training and the Senticap test split for evaluation would result in scores that provide only little information about the actual model capabilities.
We thus generate a second set of augmented COCO captions, based on the Senticap ANPs. 
These ANPs indicate which nouns and adjectives jointly appear in Senticap captions, independent of the sentiment actually expressed by the underlying image. 
Hence, to insert these sentiment adjectives into captions such that they represent the image content, we locate nouns appearing in the ANPs to sample and insert one of the adjectives corresponding to the detected nouns, utilizing the method from above.
We thus obtain two different sets of training captions: COCO Attributes-augmented captions paired with COCO Attributes and Senticap-augmented captions, which are composed of 520K captions with positive adjectives and 485K captions with negative adjectives.

\section{The Style-SeqCVAE Approach}
To obtain image descriptions with styles grounded in images, we first extract the attributes associated with objects in the visual scene. 
Equipped with these style features, we formalize our Style-SeqCVAE with a structured latent space to encode the localized image-grounded style information.

\subsection{Image semantics and style information}
Dense image features and corresponding object detections are extracted using a Faster R-CNN \cite{Girshick2015}.
In order to obtain the different object attributes, we add a dense layer with sigmoid activation for multi-label classification.
The loss term of the classification layer $L_\text{CN}$, which traditionally consists of a loss term for background/foreground class scores ($L_\text{cls}$), and a term for regression targets per anchor box ($L_\text{reg}$), is extended with another loss $L_\text{att}$:
\begin{equation} \label{eq:loss_rpn}
\begin{split}
L_\text{CN}(\{p_{i}\} ,\{b_{i}\}, \{a_{i}\}) =\frac{1}{N_\text{cls}}\sum _{i} L_\text{cls}\left( p_{i} ,p^{*}_{i}\right)&+\lambda_1 \frac{1}{N_\text{reg}}\sum _{i} p^{*}_{i} L_\text{reg}\left( b_{i} ,b^{*}_{i}\right) \\
 &+ \lambda_2\frac{1}{N_\text{att}}\sum _{i} \beta_i L_\text{att}\left( a_{i} ,a^{*}_{i}\right),
\end{split}
\end{equation}
where $a_i^*$ are ground-truth attribute annotations and $a_i$ denotes the predicted probabilities for each particular attribute category being present at anchor $i$. 
$L_\text{cls}$ is the binary cross-entropy loss between the predicted probability $p_i$ of anchor $i$ being an object and the ground-truth label $p^*_i$. 
The regression loss $L_\text{reg}$ is the smooth $L_1$ loss \cite{Girshick2015} between the predicted bounding box coordinates $b_i$ and the ground-truth coordinates $b_i^*$. 
$L_\text{att}$ is the class-balanced softmax cross-entropy loss \cite{Cui_2019_CVPR}.
$N_\text{cls}$, $N_\text{reg}$, and $N_\text{att}$ are the normalization terms.
$\lambda_1$ and $\lambda_2$ are the regularization parameters. 
Here, $\beta_i=1$ if there is an attribute associated with anchor $i$ and $\beta_i=0$ otherwise.

\begin{figure}[t]
\begin{subfigure}[b]{0.55\textwidth}
\centering
\includegraphics[width=\linewidth]{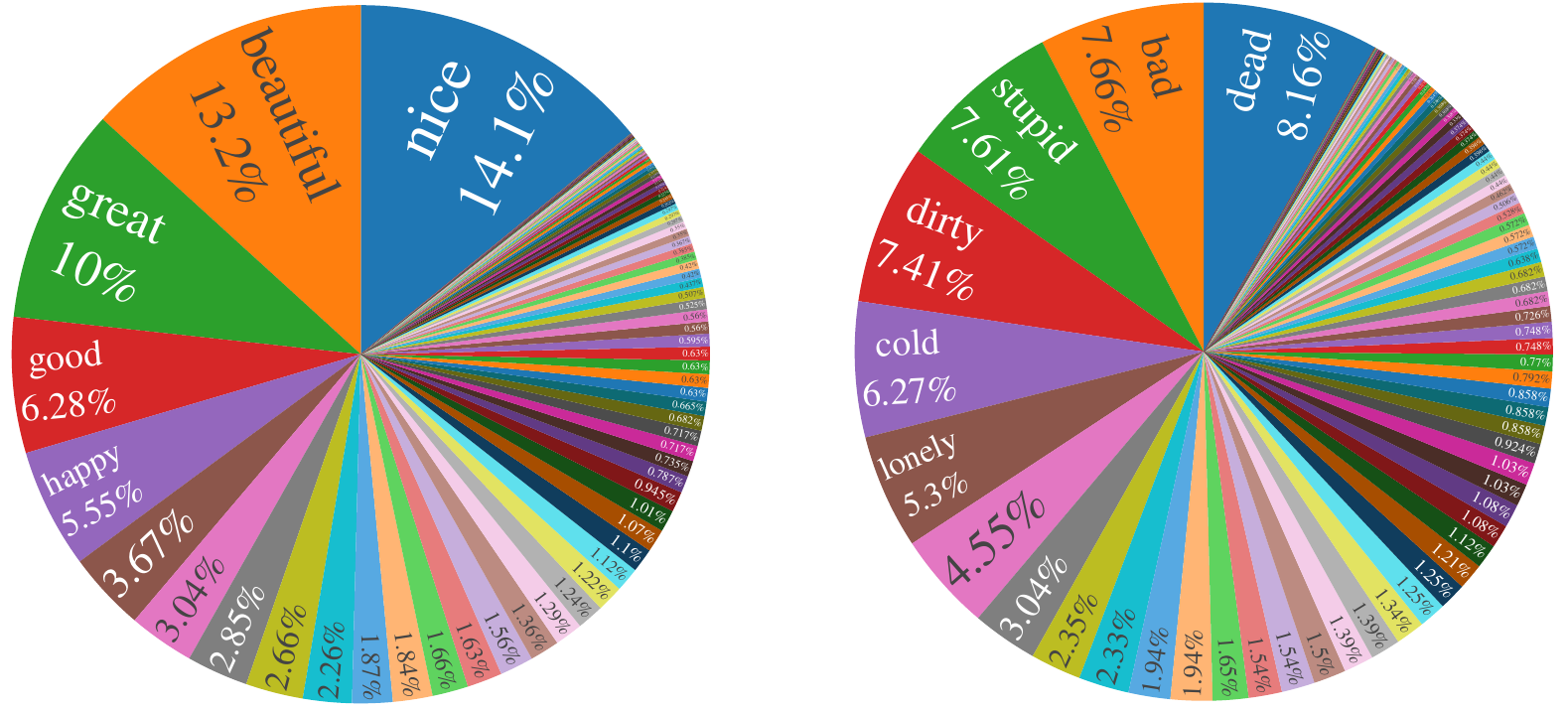}
\caption{Frequency of positive \emph{(left)} and negative \emph{(right)} captions.}
\label{fig:methods_senticap_distr}
\end{subfigure}
\hfill
\begin{subfigure}[b]{0.4\textwidth}
\centering\includegraphics[width=\linewidth]{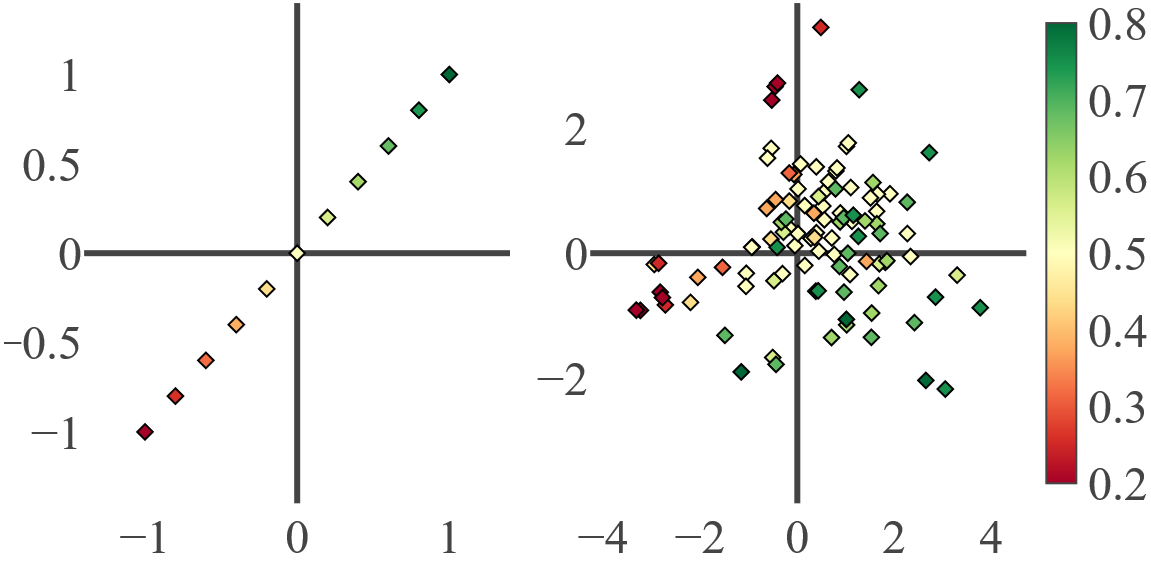}
\caption{SentiWordNet scores of detected object attributes \emph{(left)}. GloVe dimensions that encode the most sentiment information \emph{(right)}.}
\label{fig:methods_glove_latent_space}
\end{subfigure}
\caption{\emph{(a)} Adjective frequency in Senticap captions. \emph{(b)} Two different latent space structuring approaches.}
\end{figure}

\subsection{The Style-Sequential Conditional Variational Autoencoder}
\begin{figure}[t]
\begin{subfigure}[b]{0.55\textwidth}
\centering
\includegraphics[width=\linewidth]{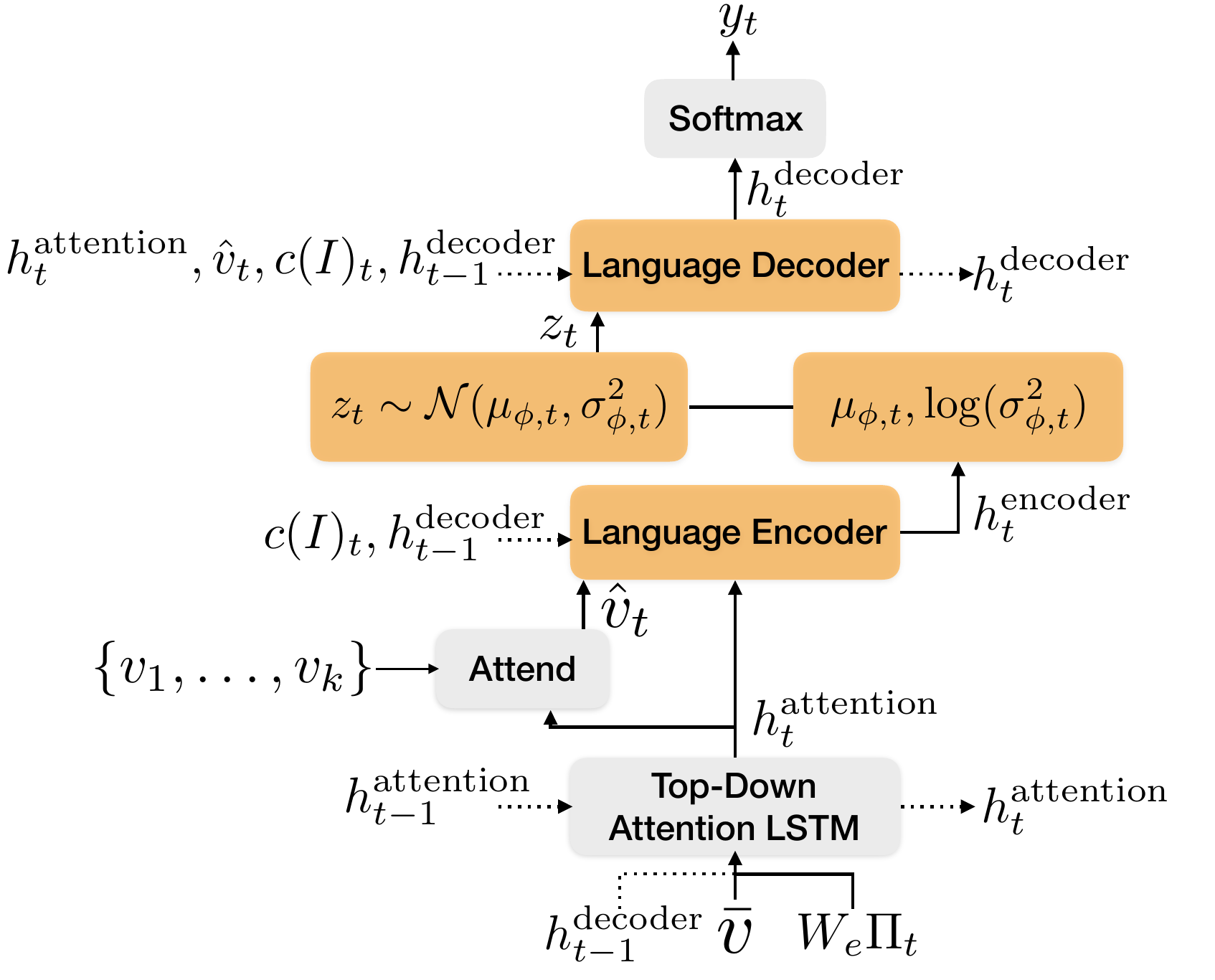}
\caption{Style-Sequential CVAE architecture.}
\label{fig:methods_model_arch}
\end{subfigure}
\hfill
\begin{subfigure}[b]{0.4\textwidth}
\centering
\includegraphics[width=\linewidth]{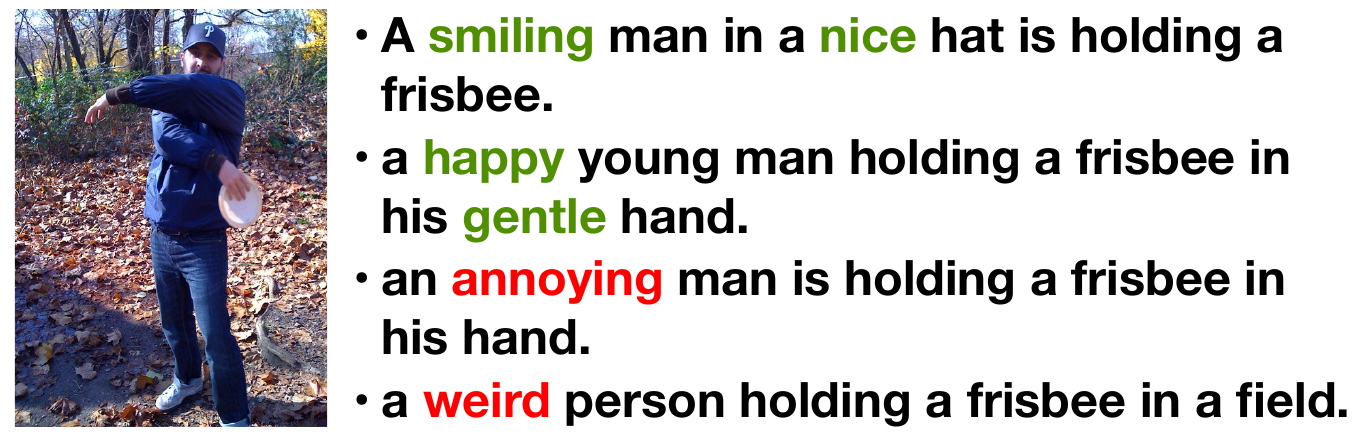}
\caption{Generated example captions with positive and negative styles.}
\label{fig:results_senticap_example}
\end{subfigure}
\caption{\emph{(a)} Style-Sequential CVAE for stylized image captioning: overview of one time step. \emph{(b)} Captions generated with Style-SeqCVAE on Senticap.}
\end{figure}

The goal of our Style-SeqCVAE framework is to generate diverse captions that reflect different perceivable style expressions in an image.
We illustrate the proposed model in \cref{fig:methods_model_arch}.
Consider an image $I$ and caption sequence $x=(x_1,\dots,x_T)$, the visual features $\{v_1,\ldots,v_K\}$ for $K$ regions of the image are extracted from a Faster R-CNN (\cf ~Eq.~\ref{eq:loss_rpn}) and the mean-pooled image features $\bar{v} =\frac{1}{K}\sum _{k} v_{k}$ are input to the attention LSTM \cite{Anderson2017}.
In this work, we propose to further encode region-level style information in $c(I)_t$ (as discussed below), and update it at each time step using the attention weights ($\alpha_t$).
This is based on the observation that the image styles can vary greatly across different regions.
To take this into account, we model a sequential VAE with explicit latent space structuring with an LSTM-based language encoder and language decoder (\cref{fig:methods_model_arch} highlights this in yellow). 
$h_{t}^{\text{attention}}$, $h_{t}^{\text{encoder}}$, and $h_{t}^{\text{decoder}}$ denote the hidden vectors of the respective LSTMs at a time step $t$.
Encoding the latent vectors $z_t$ at each time step based on the reweighted image regions and corresponding component vectors $c(I)_t$ enables the model to structure the latent space at the image region level instead of only globally.
Similar to Anderson et al.~\cite{Anderson2017}, the relevance of the input features at a specific time step $t$ depends on the generated word $W_{e} \Pi _{t}$, where $W_{e}$ is a word embedding matrix and $\Pi_t$ is the one-hot encoding of the input word. 
Given the attended image feature $\hat{v}_{t}$, the latent vectors $z$ are encoded in the hidden states $h_{t-1}^{\text{attention}}$ and $h_{t-1}^{\text{decoder}}$.
Moreover, to allow for image-specific localized style information, we enforce an attribute-based structured latent space. 

The log-evidence lower bound at time step $t$ is given by
\begin{equation}
\begin{split}
    \log p\left( x_{t} |I,x_{< t},z_{\leq t},c(I)_{t}\right) \geq \;& \mathbb{E}_{q_{\phi}}\left[\log p_{\theta }\left( x_{t} |I,x_{< t},z_{\leq t},c(I)_{t}\right)\right]\\
    &-D_{KL}[ q_{\phi }( z_{t}|I,x_{< t},z_{< t},c(I)_{t}) \,\|\, p_{\theta}( z_{t}|c(I)_{t})].
\end{split}
\end{equation}
Here, $p_\theta$ is the prior distribution parameterized by $\theta$ and $q_\phi$ denotes the variational posterior distribution with parameters $\phi$.

\myparagraph{Attribute-specific latent space structuring.} \label{sec:methods_att_latent_space}
The choice of the prior contributes significantly to how the latent space is structured. 
The additive Gaussian prior has proven to be beneficial for both diversity and controllability of the caption generation process \cite{Aneja2019,Wang2017}.
Unlike \cite{Wang2017}, where the latent space is constrained based on the objects, we instead leverage attributes to encode the styles in the image.
Specifically, available attributes are explicitly assigned to one of the image features $\{v_1,...,v_k\}$ and can thus be divided into subsets $A_{k}=\{a_{k,1},...,a_{k,j}\}$ containing $J_k$ different attributes. 
Furthermore, we adopt the attention mechanism of \cite{Anderson2018}, which provides a set of weights $\alpha_t = \{\alpha_{t,1}, ..., \alpha_{t,k}\}$ to readjust the impact of each image feature $v_k$ at every time step $t$.
Accordingly, a weight $\alpha_{t,k}$ is also mapped to the subset $A_k$, which belongs to the image feature $v_k$.
This property is exploited for the attribute-specific latent space structuring and makes it possible to weight the contribution of each individual attribute set $A_k$. We assume that the approximate style of an image region $v_t$ is represented by the total of all associated attributes $A_{k}$.
The additive Gaussian latent space can then be reformulated to calculate a $\mu_t$ at each time step $t$ as
\begin{equation}
\mu_t =\sum ^{K}_{k=1}\frac{\alpha _{t,k}}{J_k}\sum ^{J_{k}}_{i=1} \mu_{i,k}.
\end{equation}
The prior mean $\mu_t$ is thus composed of the attention-weighted average of each image region-specific linear combination of $\mu_{i,k}$. 
The variance $\sigma_t^2$ does not need to be calculated explicitly as $\sigma_i^2$ is equal for all attribute categories $i$.
However, randomly initialized, attribute category-specific Gaussian components $\mu_i$ do not reflect any semantic or contextual similarities between different attributes. 
Therefore, in this work two alternative attribute category-specific $\mu_i$ initialization approaches are pursued:
In the first case, we uniformly initialize each $\mu_i$ to the SentiWordNet score \cite{swd} corresponding to attribute category $i$.
In the following, this latent space is referred to as the \emph{SentiWordNet} latent space.
In the second case, both sentimental and semantic characteristics of different attributes are taken into account by initializing each $\mu_i$ with the $n$ dimensions of GloVe vectors corresponding to the attribute category $i$ that most strongly encode its word sentiments. These dimensions are identified by an application of PCA on the 20 COCO Attributes labels with the strongest SentiWordNet scores. If the number of extracted dimensions is less than the desired dimensionality $z$, its elements are simply repeated to upscale it. We refer to this setup as \emph{SentiGloVe} latent space. Both latent space structuring approaches are exemplified in \cref{fig:methods_glove_latent_space}.
With $\mu_i$ already encoding attribute-specific information, it serves as explicit input $c(I)_t$ to the encoder and decoder, thus $c( I)_{t} = \mu_t$. 
While training, the encoder produces Gaussian parameters $\mu_{\phi,t}$ and $\log \sigma_{\phi,t}^2$, which are used to encode a latent $z_t$ and calculate the KL-Divergence
\begin{equation}
\begin{split}
 D_{KL}[ q_{\phi }( z_{t}|I,x_{< t},z_{< t},c(I)_{t}) \,\|\, p( z_{t}|c(I)_{t})] &= \log\left(\tfrac{\sigma_t}{\sigma_{\phi,t}}\right) +\tfrac{1}{2\sigma_t ^{2}}{\mathbb{E}_{q_{\phi}}}\!\left[\left\Vert z_t-\mu_t \right\Vert ^{2}\right]\!-\!\tfrac{1}{2}\\
&= \log\left(\tfrac{\sigma_t }{\sigma _{\phi,t}}\right) +\tfrac{\sigma^{2}_{\phi,t} +\left\Vert \mu_{\phi,t} -\mu_t\right\Vert ^{2}}{2\sigma_t^{2}} -\tfrac{1}{2},
\end{split}
\end{equation}
which is used to maximize the variational lower bound. 
Furthermore, $z_t$ is provided to the decoder to produce the output word $y_t$ of the current time step.
During generation, the encoder is dropped and $z_t$ values are sampled from the same attribute-specific additive Gaussian prior that is used while training. 
The attributes attached to the image features are actual (hard) detections from the image feature extractor instead of ground-truth annotations.

\myparagraph{Sentiment-specific latent space structuring.}\label{sec:methods_senti_latent_space}
Furthermore, we extend our approach to the Senticap dataset available for stylized image captioning and thus allow for a quantitative comparison of the proposed framework with existing work. 
The Senticap dataset provides positive as well as negative captions for images in the dataset.
This implies that the captions in the Senticap dataset do not represent a variety of image-anchored styles and therefore, we cannot expect the latent space structure defined above to be particularly helpful.
Thus, we take into account the COCO captions which are augmented with Senticap adjectives and are either labeled as ``positive" or ``negative" and define a simple latent space structure around the two clusters. 
Additionally, a third cluster is defined for captions with a neutral sentiment, \eg~original COCO captions.
In contrast to the attention-weighted, attribute-based latent space structures defined above, the mean remains constant over all time steps and only depends on the provided sentiment identifier in this setup. 
And since each caption is distinctively assigned to one of these three clusters, the additive Gaussian structuring is not suitable. 
Instead, the prior mean is fully defined by one of the three predefined clusters with values $c(I)_t \in \{-0.5,0.0,0.5\}$ depending on the negative, neutral, or positive sentiment identifier.
In this case, $\mu_t = c( I)_t$ and 
$p(z_{t} |c( I)_t ) =\mathcal{N}\left( z_{t} \ \middle| \mu_t ,\sigma ^{2} I\right)$. 
Similar to the latent space structuring approaches described above, the prior variance $\sigma^2_t$ is initially set and identical for all three clusters. 
Having obtained these prior parameters and $c(I)_t$, the training procedure is identical to that of the attribute-specific latent space.
During evaluation, we generate diverse captions for a given sentiment by selecting one of the three clusters (by choosing the prior mean associated with the intended sentiment).

\myparagraph{Style-based constrained beam search.} 
Since only a fraction of COCO images are annotated with attributes, we rely on COCO Attributes-augmented captions for a fraction of training images. 
When the model is now trained using a combination of the original and attribute-augmented COCO captions, the attribute words occur rarely in the decoding procedure, especially the ones that are extremely underrepresented in the training data. 
This issue is optionally addressed by presenting the detected object attributes as constraints during the decoding procedure using constrained beam search (CBS) \cite{Anderson2016}.

\section{Experiments}
We next evaluate and analyze our approach for the generation of a diverse set of image captions for a particular image with image-grounded styles.
We evaluate on the Senticap and COCO datasets. 
On Senticap, we use the Senticap-augmented captions for training.
When evaluating on the standard COCO dataset, we utilize the COCO Attributes-augmented captions for training in order to encode image-specific style information in the latent space.

\myparagraph{Evaluation metrics.}
The accuracy of the captions is evaluated with standard metrics -- Bleu (B) 1--4 \cite{bleu}, CIDEr (C) \cite{chen2015microsoft}, ROUGE (R) \cite{rouge}, and METEOR (M) \cite{meteor}.
Similar to Mathews et al.~\cite{Mathews2015}, we consider the percentage of candidate captions containing at least one of the Senticap ANPs as part of the evaluation, referred to as SEN\%. 
Additionally, we report the precision (SP) and recall (SR) of sentiment adjectives occurring in candidate and reference captions.

\subsection{Evaluation on the Senticap dataset} \label{sec:results_senticap_eval}
\begin{table}[t]
  \centering
  \scriptsize
    \caption{Top-1 oracle scores on Senticap test split captions with positive and negative sentiments.
  $n$ denotes the number of captions generated per image.}
  \begin{tabularx}{\linewidth}{@{}Xcccccccc|ccccccc@{}}
     \toprule
            &  &\multicolumn{7}{c}{Positive} & \multicolumn{7}{c}{Negative}\\
           \cmidrule(lr){3-9} \cmidrule(lr){10-16}
     Method & $n$ & B1 & B2 & B3 & B4 & R & C & M & B1 & B2 & B3 & B4 & R & C & M\\
      
    \midrule
Senticap \cite{Mathews2015} & 1 & 49.1 & 29.1 & 17.5  & 10.8  & 36.5 & 54.4  & 16.8  & 50.0 & 31.2 & 20.3 & 13.1 & 37.9  & 61.8  & 16.8\\
StyleNet \cite{Gan2017}     & 1 & 45.3 & -- & 12.1 & -- & --& 36.3  & 12.1  & 43.7 & --  & 10.6 & --  & -- & 36.6 & 10.9  \\
You et al. \cite{You2018}   & 1 & 51.2 & 31.4 & 19.4  & 12.3  & 38.6  & 61.1 & 17.2 & 52.2 & 33.6 & 22.2 & 14.8 & 39.8 & 70.1& 17.1 \\
Chen et al. \cite{Chen2018b} & 1  & 50.5     & 30.8     & 19.1     & 12.1 & 38.0    & 60.0    & 16.6 & 50.3                            & 31.0    & 20.1 & 13.3 & 38.0  & 59.7 & 16.2    \\
Senti-Attend \cite{Nezami2018} & 1  & 57.6  & 34.2 & 20.5 & 12.7  & 45.1   & 68.6 & 18.9 & 58.6 & 35.4 & 22.3 & 14.7 & 45.7  & 71.9 & 19.0   \\
MSCap \cite{Guo2019}  & 1  & 46.9     & --   & 16.2     & --   & --  & 55.3    & 16.8 & 45.5                            & --  & 15.4  & --  & -- & 51.6  & 16.2    \\
AttendGAN \cite{Nezami2019}           & 1  & 56.9     & 33.6     & 20.3     & 12.5  & 44.3  & 61.6  &18.8& 56.2                            & 34.1 & 21.3   & 13.6  & 44.6 & 64.1 & 17.9    \\
Memcap \cite{Zhao2020}                & 1  & 51.1     & --   & 17.0     & --   & --  & 52.8    & 16.6 &49.2                   & --  & 18.1    & --  & --  & 59.4 & 15.7    \\ 
\midrule
Karayil et al. \cite{Karayil2019}  & 1 & 54.7 & 34.6 & 22.0 &  14.4 & 41.8  & 46.1 & 18.5 & 57.0 & 36.2 & 23.4 & 15.1 & 44.5  & 50.9 & 19.9 \\
& 10 & 65.6 & 43.9 & 29.5 & 20.2 & 48.8 & 63.1 & 22.1 & \bfseries 67.6 & 46.3 & 31.9 & 21.9 &  50.4 & 68.8 & 23.5 \\ 
\midrule
\multirow{2}{*}{Style-SeqCVAE} & 1 & 53.8  & 33.2 & 20.1     & 12.5 & 41.5 & 71.1 & 19.7 & 55.2 & 34.5 & 21.5 & 13.4 & 41.5 & 75.5 & 19.4                                \\
& 10 & \bfseries 66.3 & \bfseries 46.3 & \bfseries 32.2 & \bfseries 22.2 & \bfseries 51.2 & \bfseries 110.5 & \bfseries 25.2 & 66.1 & \bfseries 46.8 & \bfseries 33.3 & \bfseries 23.7 & \bfseries 50.9  & \bfseries 111.9 & \bfseries 24.4   \\
              \bottomrule
  \end{tabularx}
  \label{tab:senticap_baseline_joint}
\end{table}
In this setting, the original COCO 2017 train split is combined with the COCO captions augmented with Senticap adjectives. 
The available captions express either a negative, neutral, or positive sentiment and, therefore, the latent space is explicitly structured around three different clusters encoding the sentiment expressed in the generated captions. 
Unless otherwise stated, constrained beam search is not used for caption generation.
Since the Senticap dataset consists of positive and negative ground-truth captions for images, not related to the actual image sentiment, prior work \cite{Chen2018b,Gan2017,Guo2019,Karayil2019,Mathews2015,Nezami2018,Nezami2019,You2018} generates a positive as well as a negative caption for a given image based on the style indicator.
Therefore, in order to compare our approach on the Senticap evaluation dataset, we generate positive and negative captions for a given image based on the sentiment-specific latent space as discussed above (\cref{sec:methods_senti_latent_space}). 
The quality of the generated positive and negative captions is reported in \cref{tab:senticap_baseline_joint}.
When generating only one caption per image ($n=1$), the achieved scores are comparable to the best-performing existing work. 
When generating $n=10$ captions, the presented approach performs clearly better than the only related work \cite{Karayil2019} that generates diverse captions conditioned on the style indicator.
This implies that unlike our Style-SeqCVAE approach, \cite{Karayil2019} does not encode as many variations in style content for a given image.
The fact that we obtain high scores with our approach on metrics that take longer $n$-grams into account indicates that appropriate adjectives related to style are inserted into the captions in a suitable place.
This also suggests that our proposed caption augmentation approach preserves the syntactic correctness of the resulting training captions. 
Most striking is the significant performance increase in the CIDEr score, which particularly rewards the use of $n$-grams that only rarely appear in the reference captions. 
This implies that explicitly structuring the latent space around fixed style-based clusters encourages even underrepresented style adjectives to prevail during decoding, especially when multiple captions are generated for an image.

In Table \ref{tab:results_sentiment_metrics}, we show that the diversity in the generated captions is the result of the different ways of expressing sentiment (and not solely based on diversity from the factual descriptions).
In this setting, we include the Senticap training set without decoding constraints (COCO + Senticap-augm) and with decoding constraints (COCO + Senticap-augm + CBS).
The proportion of reference sentiment adjectives appearing in the candidate captions doubles from $\sim$0.15 to $\sim$0.3 when producing 10 captions per image, which shows clearly that the diversity in the captions also has an effect on sentiment expression. Captions generated by \cite{Mathews2015} exclusively consider the most dominant adjectives in the training/test captions without much diversity. Thus, its slightly higher SP score is expected, given the massive adjective imbalance of Senticap.
The Senticap-augmented model (COCO + Senticap-augm) inserts at least one ANP with matching sentiment in almost every caption $(>93.6\%)$.
Furthermore, we do not observe a significant improvement in any metric when attribute-based CBS constraints are applied.
This shows that the latent space of our Style-SeqCVAE can effectively model the style information in the latent space.
Additionally, the high CIDEr score supports that the captions generated by our approach are accurate.
Qualitative examples in \cref{fig:results_senticap_example} show the captions with varied styles generated with Style-SeqCVAE.
For the given image, the diverse captions reflect the positive as well as negative sentiments showing that the latent space of our approach effectively captures style information of the data distribution.
\begin{table}[t]
\scriptsize
\centering
\caption{Additional evaluation on the Senticap test set. SP denotes the sentiment precision and SR the sentiment recall.}
\begin{tabularx}{\textwidth}{cXrrrrrrrrrr}
\toprule
\textbf{n} &\textbf{Method} & \textbf{B1} & \textbf{B2} & \textbf{B3} & \textbf{B4} & \textbf{R} & \textbf{C} & \textbf{M} & \textbf{\%SEN} & \textbf{SP} & \textbf{SR} \\ 
\midrule
\multirow{3}{*}{1}  & Senticap,   Mathews et al. \cite{Mathews2015}         & 48.8  & 29.8 & 18.7  & 11.8  & 37.2  & 56.6  & 16.8 & 87.5 & \bfseries 0.33 & 0.14 \\
& COCO + Senticap-augm.  & 54.5  & 33.9 & 20.8 & 13.0 & 41.5 & 73.3 & 19.6 & 93.6 & 0.30 & 0.15 \\
& COCO + Senticap-augm. + CBS & 54.7 & 34.0  & 21.1 & 13.2  & 41.7& 74.1 & 19.7& 100.0  & 0.30  & 0.15\\
\midrule
\multirow{2}{*}{10} & COCO + Senticap-augm.  & 66.2  & 46.6   & 32.8  & 23.0  & 51.0  & 111.2  & 24.8  & 99.3 & 0.26 & 0.30 \\
 & COCO + Senticap-augm. + CBS & \bfseries 66.3 & \bfseries 47.2 & \bfseries 33.6 & \bfseries 23.9 & \bfseries 51.5 & \bfseries114.5 & \bfseries 25.3 & \bfseries 100.0 & 0.25 & \bfseries0.32\\
\bottomrule
\end{tabularx}
\label{tab:results_sentiment_metrics}
\vspace{-0.5em}
\end{table}

\subsection{Evaluation on the COCO dataset}

We now show that our approach along with the extended COCO Attributes-augmented captions can generate diverse captions with styles anchored in the image.
Since the COCO test split contains only descriptions of the scene composition, it is difficult to quantitatively evaluate for stylized caption generation on the COCO dataset. 
Therefore, we first present a qualitative analysis of the Style-SeqCVAE approach.
For this, we consider captions generated directly from the latent space (without CBS) as well as with CBS constraints to account for the rarity of the attributes in the dataset.
Unlike \cite{Anderson2016}, which presents class labels as constraints, in this work, we enforce style information from automatically extracted image attributes as constraint to the caption generator.
For an effective application of CBS constraints in conjunction with Style-SeqCVAE, two different decoding strategies are considered: In weak constraining, the constraint is to use at least one attribute from the set of detected attributes during decoding for a given image.
The detected attributes are obtained directly from the image extractor (Faster R-CNN) trained using \cref{eq:loss_rpn}.
In individual constraining, the attribute to be inserted is explicitly selected from the set of detected attributes for an image and provided as a constraint.
Since the model is forced to take this attribute into account, it enables to also consider attributes that are underrepresented in the training data and which might otherwise be ignored.
When generating multiple captions for a given image, a constraint is randomly selected from the set of detected attributes for each of the captions.
This encourages diversity in the style of the generated captions for each image.

\begin{figure}[t]
\centering
\includegraphics[width=\textwidth]{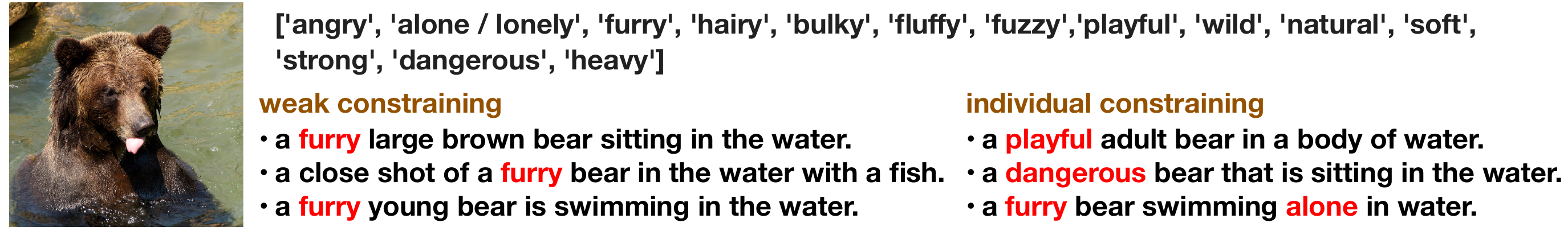}
\vspace{-5mm}
\caption{Examples of captions generated by either constraining the decoding procedure on the whole set of detected image attributes (weak constraining) vs. one specific attribute as constraint per caption (individual constraining).}
\label{fig:results_constraining_comparison}
\end{figure}
\begin{figure}[t]
\vspace{-0.6em}
\includegraphics[width=\textwidth]{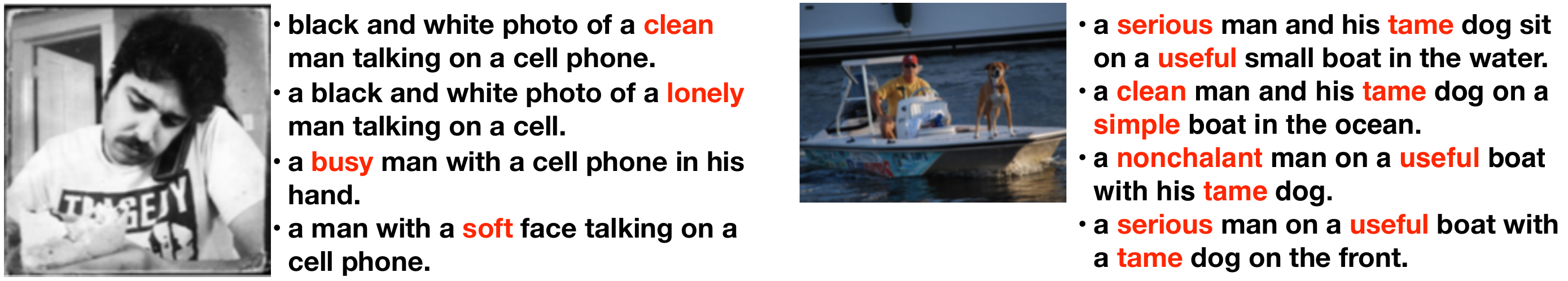}
\vspace{-5mm}
\caption{Generated example captions with SentiGloVe-structured latent space and individual decoding constraints for images with multiple objects.}
\label{fig:results_att_vae_example1}
\end{figure}

With the weak constraining mechanism, we observe that attributes that occur in high frequency in the training data are repeated in the generated captions. 
As shown in \cref{fig:results_constraining_comparison}, the detected attribute ``furry'' associated with the object ``bear'' occurs across all the generated captions for the given image.
In case of individual constraining, where an attribute is randomly provided as constraint from the set of detected attributes, this effect is not present.
For example, in \cref{fig:results_constraining_comparison}, we observe diversity in style with attributes like ``playful'' and ``dangerous'' describing the object ``bear''.
In \cref{fig:results_att_vae_example1}, we show an extension of the individual constraining procedure to occurrences of multiple objects in an image.
Here, the model is forced to insert an attribute for at least two detected objects.
For example, in \cref{fig:results_att_vae_example1} the diverse attributes are successfully inserted for the objects ``man'' and ``face'' or for ``man'', ``dog'', and ``boat'' in the captions of the respective images.

The quantitative evaluation of the proposed framework on the COCO dataset is limited due to the lack of ground-truth captions for direct comparison purposes (see appendix).
To obtain a better assessment in spite of that, we compare our method with various established approaches for diverse image captioning. 
Here, we use the SentiGloVe latent space to generate diverse captions.
Following the standard evaluation protocol of \cite{Aneja2019,mahajan2020latent,Wang2017}, in \cref{tab:results_fact_caps_acc} we show the top-1 oracle performance using 20 samples on various metrics for caption evaluation.
We observe generally competitive performance, especially in the case where a standard deviation of 2 is used when sampling from the latent space. 
This is despite the fact that, unlike previous work on diverse image captioning, our model focuses on stylistic diversity, which is not represented in the ground-truth captions of the test set.
The high accuracy scores, moreover, demonstrate the ability of the approach to successfully model the attribute-based style information to generate semantically coherent captions (\cf~\cref{fig:results_constraining_comparison}).
\begin{table}[t]
\scriptsize
\begin{minipage}{.59\textwidth}
\caption{Evaluation of semantic accuracy.}
\begin{tabularx}{\textwidth}{@{}Xccccccccc@{}}
\toprule
    \bfseries Method  &  \bfseries std &  \bfseries B1 &  \bfseries B2 &  \bfseries B3 &  \bfseries B4 &  \bfseries R &  \bfseries C &  \bfseries M \\ \midrule
 Div-BS \cite{vijayakumar2016diverse} & -- & 83.7  & 68.7  & 53.8 & 38.3 & 65.3 & 140.5 &  35.7  \\
 AG-CVAE \cite{Wang2017} & -- & 83.4 & 69.8  & 57.3 & 47.1 & 63.8 & 125.9 & 30.9 \\
 POS \cite{deshpande2019fast} & -- & \bfseries 87.4  & \bfseries73.7 & \bfseries 59.3  & 44.9  & \bfseries 67.8 & \bfseries 146.8 & \bfseries 36.5 \\
 Seq-CVAE \cite{Aneja2019}& -- & 87.0 & 72.7 & 59.1 & 44.5 & 67.1 & 144.8 & 35.6\\ \midrule
 \multirow{2}{*}{Style-SeqCVAE} & 1 & 84.2 & 69.5 & 56.0 & 44.7 & 63.4 & 130.4 & 31.2\\
 & 2 & 86.6  & 72.0  & 58.8 & \bfseries 47.6 & 65.9 & 137.2 & 32.8 \\ 
\bottomrule
\end{tabularx}
\label{tab:results_fact_caps_acc}
  \end{minipage}
  \hfill
   \begin{minipage}{.38\textwidth}
\caption{Caption diversity.}
\begin{tabularx}{\textwidth}{@{}Xccc@{}}
\toprule
 \textbf{Method} & \textbf{std} &  \bfseries Div-1 & \bfseries Div-2 \\  
 \midrule
 Div-BS \cite{vijayakumar2016diverse} & --  & 0.20  & 0.26 \\
 AG-CVAE \cite{Wang2017}  & --  & 0.24 & 0.34 \\
 POS \cite{deshpande2019fast}  & --   & 0.24   & 0.35\\
 Seq-CVAE \cite{Aneja2019}    & --   & 0.25 & \textbf{0.54} \\ 
 \midrule
 \multirow{2}{*}{{Style-SeqCVAE}} & 1 & 0.24 & 0.31\\
 & 2  & \textbf{0.29}    & 0.43   \\ 
\bottomrule
\end{tabularx}
\label{tab:results_fact_caps_div}
  \end{minipage}
\end{table}

Furthermore, we quantitatively compare our approach against \cite{Aneja2019,deshpande2019fast,vijayakumar2016diverse,Wang2017} for diversity.
In \cref{tab:results_fact_caps_div}, we use Div-1 and Div-2 for evaluation, where Div-\textit{n} is the ratio of distinct $n$-grams per caption to the total number of words generated per set of diverse captions.
Following prior work, the scores are based on the top-5 captions with highest CIDEr scores of the COCO validation split \cite{KarpathyF17} and consensus re-ranking \cite{consus_reranking} is applied before selecting the top-5 captions.
Owing to the unconstrained latent space, Seq-CVAE \cite{Aneja2019} generates captions with high diversity.
The scores on the diversity metrics indicate that in comparison to other approaches that also impose constraints in the latent space, \eg~AG-CVAE\cite{Wang2017}, our attribute-based latent space exhibits a higher diversity in style.
This can be attributed to the structured sequential latent space, where the distribution of attributes is better captured conditioned on the image.
This highlights the advantages of our style-specific latent space for diverse caption generation.

\section{Conclusion}
We present Style-SeqCVAE, a variational autoencoder framework to encode localized style information representative of the visual scene.
The structured latent space exploits the object attributes from the associated images to model the characteristic styles.
In particular, we leverage the attribute information in different image regions to express different styles present in the image.
The key to the success of the proposed latent space is the combination of attribute-based style information and region-based image features via an attention mechanism for coherent caption generation. 
Our experiments demonstrate that our approach generates diverse and accurate captions with varied styles expressed in the image.

\vspace{1em}
{\small
\noindent\textbf{Acknowledgement}.
This project has received funding from the European Research Council (ERC) under the European Union’s Horizon 2020 research and innovation programme (grant agreement No.~866008).}

\bibliographystyle{splncs04}
\bibliography{egbib}

\clearpage
\appendix

\section{Implementation Details}
In this appendix, we provide the architectural details of our Style-SeqCVAE model and additionally describe several important implementation details. 

\subsection{Image feature extraction}
For training, the Faster R-CNN is initialized with ResNet-101 \cite{He2015} weights and pretrained for classification on ImageNet \cite{imagenet}. 
All images are rescaled such that the pixel size for the shortest side is 600. 
During training, the model parameters are optimized for 10 epochs with 12 images per batch and 256 region proposals per image. 
The weights are updated using a simple stochastic gradient descent optimizer with momentum ($\beta=0.9$), setting the initial learning rate to 0.001. 
A learning rate decay of 0.1 is applied every 4 epochs.

Following \cite{Anderson2017}, the 2048-dimensional average-pooled activations of the fc7-layer are extracted as image region features. 
Between 10 and 100 features are extracted per image, depending on whether the probability of the detected object class exceeds a confidence threshold of 0.5. 
The threshold for extracted attribute detections is set to 0.3, which are extracted for object classes that also appear in the COCO Attributes annotations.

\subsection{Image captioning hyperparameters} 
In general, the latent space of the Style-SeqCVAE is trained using the COCO 2017 train split. 
The hidden size of the encoder, decoder, and attention LSTM is set to 900; the latent space has 150 dimensions. 
For initialization of all experiments with the SentiGloVe latent space structure, the 10 GloVe dimensions that encode the most sentiment information are extracted using PCA, which are scaled to the dimensionality of 150.
Regardless of the selected prior distribution, its standard deviation during training is always set to 1, and also at test time if not explicitly specified otherwise. 
A stochastic gradient descent optimizer with momentum ($\beta=0.9$) and initial learning rate 0.015 is applied to update the weights. 
The gradients are clipped to 12.5 to avoid exploding gradients. 
The model is trained for 70K iterations with 150 image-caption pairs per batch. 
For the sake of comparability, the beam size of CBS is set to 1.

\section{Additional Quantitative Evaluation}

\begin{figure}[t]
\centering\includegraphics[width=\linewidth]{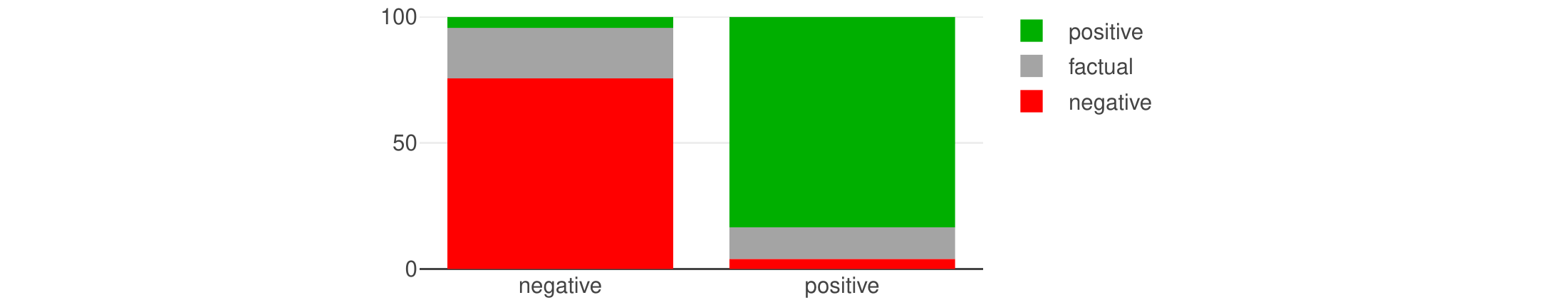}
\caption{Classification results by applying separately trained text sentiment classifiers (see description below) to generated captions with intended negative sentiment \emph{(left)} and positive sentiment \emph{(right)}. Each bar indicates the proportion of captions of the respective subsets that are classified as negative, neutral, or positive.}
\label{fig:results_senticap_sentiments}
\end{figure}

\paragraph{Senticap dataset.} In order to obtain at least an approximate quantitative assessment of the style in the generated captions, text sentiment classifiers can be utilized to estimate the expressed style of both the stylized candidate captions and corresponding factual COCO reference captions. 
This gives some insights into whether style information is actually added to generated captions and allows to compare the sentiment of candidate captions with the sentiment of COCO Attributes ground-truth annotations. 
To this end, three different text sentiment analysis approaches are considered. 
The first and simplest one just retrieves polarities from the opinion lexicon \cite{liu_hu_lexicon_senti_analyzer} for each word in a sentence and measures which polarity is most present in a caption. 
The second text analyzer, Vader, is a parsimonious rule-based model, originally developed to analyze the sentiment of social media text \cite{vader_senti_analyzer}. 
And finally, a much more sophisticated recurrent neural tensor network is applied, which is trained on the Sentiment Treebank of Stanford NLP \cite{stanfordNLP_senti_analyzer}.
The sentiment for each particular caption is determined by majority vote. 
The accumulated sentiment assignments for both subsets are visualized in \cref{fig:results_senticap_sentiments}, which clearly shows that the sentiment analyzers assign the overwhelming majority of captions to the intended sentiment class. 
This suggests that using our approach for inserting adjectives with strong sentiments is an effective tool for influencing the perceptible style in the text. 

\begin{figure}[!htb]
{\centering\includegraphics[width=\linewidth]{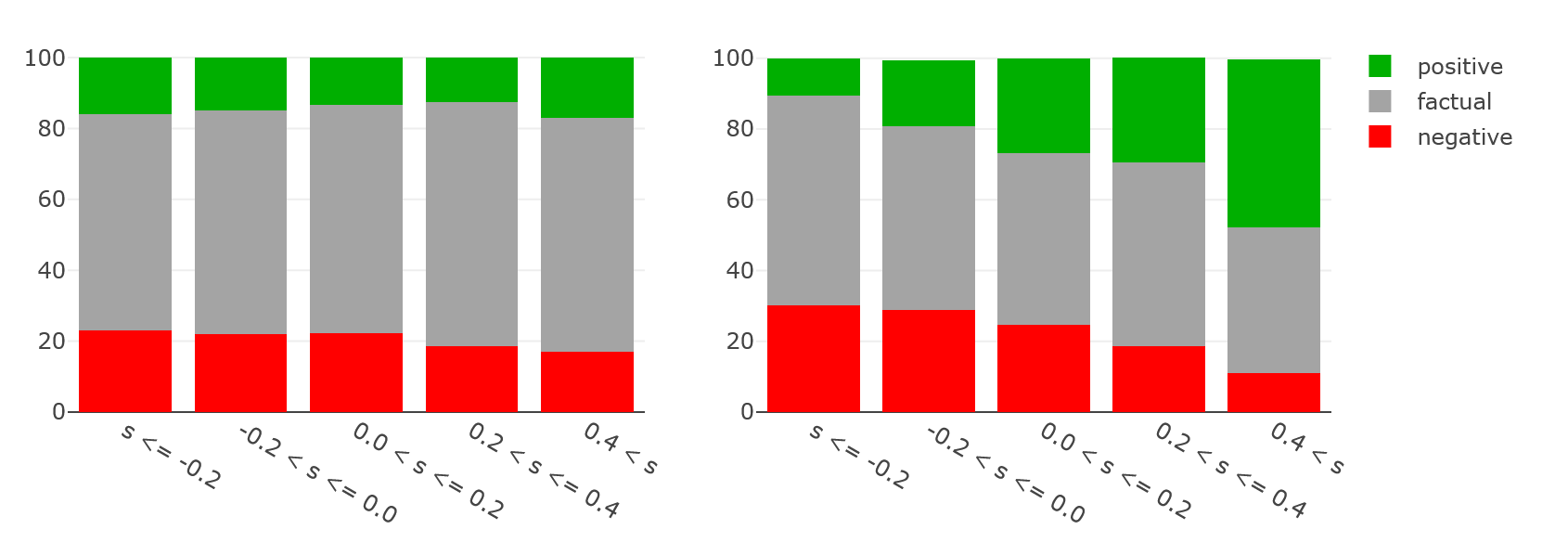}}
\caption{Text classification results by applying the text sentiment classifiers to original COCO reference captions \emph{(left)} and COCO Attributes-augmented reference captions \emph{(right)}. Each bar indicates the proportions in which the captions of the respective subsets are classified as negative, neutral, or positive.} 
\label{fig:gt_sentiments}
\end{figure}

\paragraph{COCO dataset.} 
For this evaluation, we divide the SentiWordNet score range of -1 to 1 into 5 intervals. 
For all images in the COCO validation set with available COCO Attributes ground-truth annotations, the average SentiWordNet score of all attributes per image is calculated and for each of the 5 intervals, 250 images are sampled with the appropriate average SentiWordNet score, which are then analyzed by the text sentiment analyzers.\footnote{Text sentiment analyzers aim to determine the overall sentiment of a piece of text, \ie~`positive', `neutral', or `negative'. This is clearly a much more coarse-grained notion than the diverse styles from COCO Attributes that we consider here. Nevertheless, this can serve as a useful quantitative assessment.} 
The results are illustrated in \cref{fig:gt_sentiments}. 
It can be clearly observed that the image styles determined on the basis of the COCO Attributes ground-truth annotations are reflected in the COCO Attributes-augmented reference captions.
We observe that in case of the original, factual COCO captions, which only describe the scene composition, the SentiWordNet scores exhibit lower correlation with the overall sentiment of the text while in case of the COCO Attributes-augmented captions, the SentiWordNet score has a stronger correlation with the overall sentiment of the caption description.
\begin{figure}[!htb]
\centering\includegraphics[width=\linewidth]{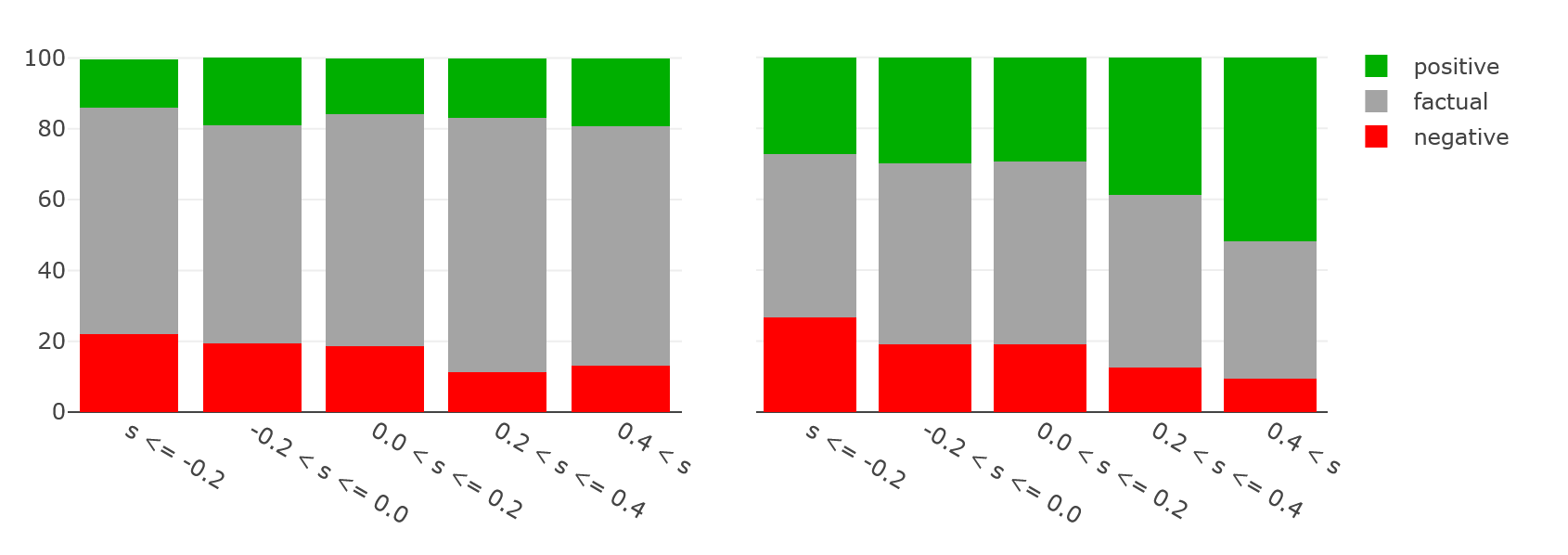}
\caption{Text classification results by applying the text sentiment classifiers to produced factual captions \emph{(left)} and produced captions with inserted adjectives \emph{(right)} from our Style-SeqCVAE. Each bar indicates the proportions in which the captions of the respective subsets are classified as negative, neutral, or positive.}
\label{fig:preds_sentiments}
\end{figure}

For the captions generated with the SentiGloVe latent space structuring, the styles are also evaluated using the text sentiment analyzers and visualised in \cref{fig:preds_sentiments}.
The text classification results of the captions generated with our Style-SeqCVAE trained on COCO Attributes-augmented captions are very similar to that of the COCO Attributes-augmented ground-truth captions.
This shows that, as expected, the sentiment distribution of the generated captions is very similar to the sentiment distribution on the ground-truth captions.
The generated captions with the model trained on both the original COCO captions as well as the COCO Attributes-augmented captions show a similar trend as for the ground-truth captions, where again the generated COCO captions have lower correlation between the SentiWordNet score and the sentiment of the overall caption. 
The generated captions based on training with COCO Attributes-augmented captions, on the other hand, have a stronger correlation, as we expect.
This suggests that the style as captured by sentiments encoded in the latent space of our Style-SeqCVAE is representative of the style present in the dataset.

\section{Generated Example Captions}
\subsection{Training Style-SeqCVAE on Senticap-augmented captions}

In \cref{tab:supp_senticap_augmented}, we provide additional qualitative examples of the stylized captions generated with our Style-SeqCVAE approach when trained with the Senticap-augmented captions.
The generated captions clearly reflect the positive and negative sentiments in the image.
\begin{table}[h]
\centering
\caption{Qualititative examples of Senticap-augmented captions with Style-SeqCVAE.}
\smallskip
\scriptsize
\begin{tabularx}{\linewidth}{@{}lX@{}}
\toprule
Image & Generated Captions \\
\midrule
 \raisebox{-51pt}{\includegraphics[width=0.16\linewidth]{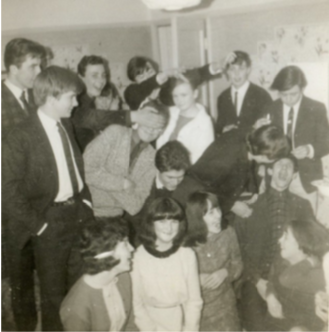}}
 & \vspace{-1.5mm}\begin{itemize}[nosep,leftmargin=*,label={\textbullet}]
\item a great group of young children are dressed in dresses and ties
\item a great group of good young cute girls in a classroom
\item a great group of excited young children in a nice room
\item a black and white photo of a great group of happy children in a nice room
\item a great group of excited people in a room with their arms
\item an ugly group of children in formal attire posing for a terrible picture
\item an old black and white photograph of weird people in a room
\item a black and white photograph of a bad old photo of a group of sad people in a room
\item an ugly group of ill children are standing in a room
\item an ugly group of sad people in a room with their arms hands
\end{itemize} 
\\[-7 pt]
\midrule
 \raisebox{-51pt}{\includegraphics[width=0.16\linewidth]{Images/supp_2_senticap.png}}
& \vspace{-1.5mm}\begin{itemize}[nosep,leftmargin=*,label={\textbullet}]
\item a new truck with a nice group of wonderful people riding on it
\item a new white truck with a nice bunch people riding on it
\item a cool truck with a bunch of people riding on back of it
\item a new white truck with a bunch of awesome people riding on it
\item a new truck with a nice bunch of people riding on the back of it
\item a new truck with a great on the back of it is driving down a sunny road
\item a white truck with dead on it parked in the middle of a lonely town street 
\item a white truck truck with a bunch of sad people riding on the back
\item an old white truck with a bunch of sad people riding on it
\item a white truck with a dead driver and riding on
\end{itemize} 
\\[-7 pt]

\midrule
 \raisebox{-51pt}{\includegraphics[width=0.16\linewidth]{Images/supp_3_senticap.png}}
& \vspace{-1.5mm}\begin{itemize}[nosep,leftmargin=*,label={\textbullet}]
\item a nice street with a great traffic light and a beautiful building 
\item a great city street with a great traffic light and a nice traffic 
\item a calm street with traffic driving down a pleasant street
\item a calm street corner with a traffic light and a beautiful traffic light
\item a traffic light a beautiful street and a beautiful building
\item a traffic light is on a lonely city street
\item a dead busy street with a bad traffic light and a damaged building building
\item a lonely street with a bad light of a dead street in the background
\item a dead street with a bad traffic light and a bad light
\item a lonely street with a bad traffic light and a damaged building in the background
\end{itemize} 
\\[-7 pt]
\midrule
 \raisebox{-51pt}{\includegraphics[width=0.16\linewidth]{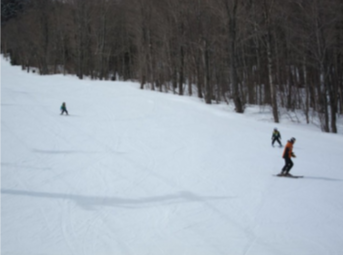}}
& \vspace{-1.5mm}\begin{itemize}[nosep,leftmargin=*,label={\textbullet}]
\item a nice couple of skiers are skiing down a pleasant hill
\item a good person skier skiing down a pleasant hill in the beautiful woods
\item a nice person skiing down a beautiful snow covered slope
\item a nice person skiing down a snowy slope in the beautiful woods
\item a nice person skiing down a snowy path in the woods
\item a weird person riding down a dirty snow covered slope
\item a weird person skiing down a rough snowy hill
\item a couple of sad people are skiing down a rough hill
\item a weird person skiing down a rough slope in the rotten woods
\item a weird person skiing down a rough hill in a forest
\end{itemize} 
\\[-7 pt]
\bottomrule
\end{tabularx}
\label{tab:supp_senticap_augmented}
\end{table}

\subsection{Training Style-SeqCVAE on COCO Attributes-augmented captions}
Additional qualitative examples of the captions generated with COCO Attributes are provided in \cref{tab:supp_attribute_augmented,tab:supp_attribute_augmented_2}. 
The example captions show that the Style-SeqCVAE generates captions representative of the image style.
\begin{table}[h]
\centering
\caption{Qualititative examples of COCO Attributes-augmented captions with Style-SeqCVAE.}
\smallskip
\scriptsize
\begin{tabularx}{\linewidth}{@{}lX@{}}
\toprule
Image & Generated Captions \\
\midrule
 \raisebox{-51pt}{\includegraphics[width=0.16\linewidth]{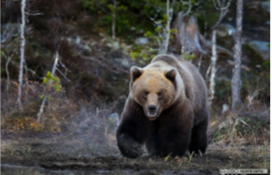}}
 & \vspace{-1.5mm}\begin{itemize}[nosep,leftmargin=*,label={\textbullet}]
\item a dangerous big brown bear sitting in the middle of a forest 
\item a calm bear that is standing in the grass
\item a bulky brown bear sitting in the middle of a forest
\item a heavy large brown bear standing in the middle of a forest 
\item a lonely bear that is sitting in the grass
\item a hairy bear that is sitting in the grass
\item a wild bear that is sitting down a tree
\item a strong big brown and black bear in the forest 
\item a furry large black bear in the middle of a forest 
\item a calm brown bear in the woods and a tree
\end{itemize} 
\\[-7 pt]
\midrule
 \raisebox{-51pt}{\includegraphics[width=0.16\linewidth]{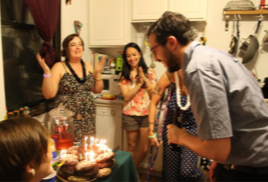}}
& \vspace{-1.5mm}\begin{itemize}[nosep,leftmargin=*,label={\textbullet}]
\item a man is celebrating with a woman in front of a yummy cake
\item a friendly man and a nonchalant woman standing over a cake
\item a group of busy people standing around a table topped with food
\item a group of engaged people standing around a table
\item a group of joyful people standing around a table
\item a nonchalant man and a nonchalant woman standing in a kitchen cutting cake 
\item a enthusiastic of people in a kitchen kitchen with a cake
\item a group of people standing around a delicious cake
\item a group of excited people standing around a table
\item a group of happy people standing around a table

\end{itemize} 
\\[-7 pt]

\midrule
 \raisebox{-51pt}{\includegraphics[width=0.16\linewidth]{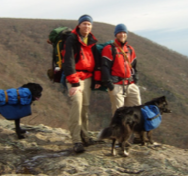}}
& \vspace{-1.5mm}\begin{itemize}[nosep,leftmargin=*,label={\textbullet}]
\item a friendly man and casual woman pose on a snowy view of a gentle dog in a picnic bag
\item a man and a woman standing on a hill with gentle dogs
\item a group of people standing next to a furry dog on a leash
\item a man in a red jacket and a backpack standing on a trail with a fleecy dog and a backpack 
\item a group of glad people standing around a dog on a leash
\item a group of people standing on a rocky top a hill with a healthy dog
\item a group of engaged people standing around a group of fleecy dogs
\item a joyful couple people are walking with their pets
\item a group of sports-minded people standing on a rocky area with a tame dog 
\item a group of people standing on a rocky of mountains with two tame dogs
\end{itemize} 
\\[-7 pt]
\midrule
 \raisebox{-51pt}{\includegraphics[width=0.16\linewidth]{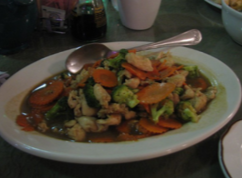}}
& \vspace{-1.5mm}\begin{itemize}[nosep,leftmargin=*,label={\textbullet}]
\item a plate of food with saucy broccoli and meat
\item a plate of food with tasty broccoli and meat on a table
\item a delicious couple of plates of food on a table
\item a plate of food with fresh vegetables and a side of background 
\item a plate of food with healthy broccoli and a soda
\item a plate of food with tasty broccoli and other
\item a plate of food with fresh vegetables and a glass
\item a plate with a salad of vegetables and soft broccoli and a glass 
\item a plate of fresh vegetables and meat on top of a table
\item a plate of food with tender broccoli and a side of a soda
\end{itemize} 
\\[-7 pt]
\bottomrule
\end{tabularx}
\label{tab:supp_attribute_augmented}
\end{table}

\begin{table}[b]
\centering
\caption{Qualititative examples of COCO Attributes-augmented captions with Style-SeqCVAE.}
\smallskip
\scriptsize
\begin{tabularx}{\linewidth}{@{}lX@{}}
\toprule
Image & Generated Captions \\
\midrule
 \raisebox{-51pt}{\includegraphics[width=0.16\linewidth]{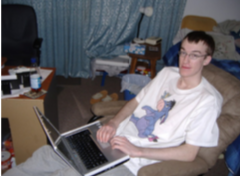}}
 & \vspace{-1.5mm}\begin{itemize}[nosep,leftmargin=*,label={\textbullet}]
\item a snugly man sitting on a chair with a laptop computer in his hand 
\item a friendly man with a laptop computer on a chair
\item a serious man sitting on a chair using a laptop computer
\item a lonely man sitting on a couch with a laptop computer
\item a peaceful man sitting on a couch with a laptop computer on his lap
\item a snugly man sitting on a couch with a laptop computer on his lap
\item a young man sitting on a chair using his laptop computer
\item a nonchalant man is sitting on a couch with a laptop computer on his lap
\item a young man sitting in a chair using his laptop
\item a joyful man sitting in a chair using a laptop computer
\end{itemize} 
\\[-7 pt]
\midrule
 \raisebox{-51pt}{\includegraphics[width=0.16\linewidth]{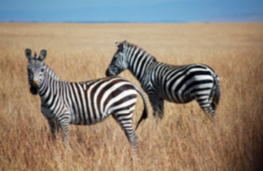}}
& \vspace{-1.5mm}\begin{itemize}[nosep,leftmargin=*,label={\textbullet}]
\item two exotic zebras are standing together in the field of dry grass 
\item a peaceful zebra and a zebra are standing in the field
\item a wild zebra and zebra in a field with a blue and light
\item two peaceful zebras are standing in the tall dry grass
\item two exotic zebras are standing in a field field with a field in the background of the mountains 
\item a peaceful zebra and a zebra in the wild
\item a hairy zebra and zebra in a field with a fence
\item a striped zebra and zebra nuzzle each another of brown grass
\item a peaceful zebra is lone zebra in the field of dry grass grass
\item a smooth zebra and a zebra are standing in the field of a field of grass
\end{itemize} 
\\[-7 pt]

\midrule
 \raisebox{-51pt}{\includegraphics[width=0.16\linewidth]{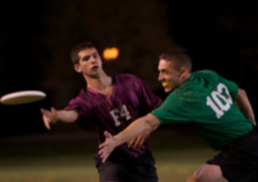}}
& \vspace{-1.5mm}\begin{itemize}[nosep,leftmargin=*,label={\textbullet}]
\item two playful guys to catch frisbee in a dark room
\item a engaged man and a nonchalant man playing a game of frisbee
\item a engaged man throwing on a frisbee while a man in the background leans on a a long 
\item a man and a nonchalant woman playing frisbee together
\item two sports-minded men are playing frisbee on a dark
\item a playful man and a nonchalant woman are throwing out a frisbee
\item a couple of sports-minded people are playing with frisbees
\item a competitive man is throwing and a frisbee in the dark
\item two sports-minded men playing frisbee in a dark dark room
\item a sports-minded man in a red shirt is throwing on a frisbee
\end{itemize} 
\\[-7 pt]
\midrule
 \raisebox{-51pt}{\includegraphics[width=0.16\linewidth]{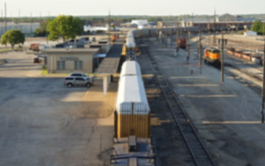}}
& \vspace{-1.5mm}\begin{itemize}[nosep,leftmargin=*,label={\textbullet}]
\item a useful train is parked on the side of the road 
\item a electric train on the tracks in a city
\item a useful train yard with a lot of cars and trucks 
\item a train yard with a truck and a useful truck
\item a dry train on a track in front of a city
\item a parked train on a track that is surrounded by buildings 
\item a train yard with a train and a truck parked on the side
\item a parked train on a track in the middle of a city
\item a useful train is parked on the side of a road
\item a useful train is parked along the street next to the beach
\end{itemize} 
\\[-7 pt]
\bottomrule
\end{tabularx}
\label{tab:supp_attribute_augmented_2}
\end{table}


\end{document}